\documentclass[runningheads]{llncs}
\usepackage{CJKutf8}
\usepackage{enumitem}
\usepackage{amsmath}
\usepackage{booktabs}
\usepackage{url}
\usepackage{hyperref}
\usepackage{amsfonts,amssymb} 
\hypersetup{
    colorlinks=true,
    linkcolor=blue,
    filecolor=blue,      
    urlcolor=blue,
    citecolor=blue,
}

\usepackage{graphicx}
\newcommand{\etal}{et al.}

\begin{document}
\bibliographystyle{splncs04}
\title{Exploiting Word Semantics to Enrich Character Representations of Chinese Pre-trained Models}
\titlerunning{Exploiting WS to Enrich Character Representations of Chinese PTMs}

\author{Wenbiao Li\inst{1,2} \and Rui Sun\inst{1,2} \and Yunfang Wu\inst{1,3}}

\authorrunning{W. Li et al.}
\institute{MOE Key Laboratory of Computational Linguistics, Peking University \and
School of Software \& Microelectronics, Peking University, Beijing, China \and
School of Computer Science, Peking University, Beijing, China\\
\email{\{2001210322,sunrui0720\}@stu.pku.edu.cn, wuyf@pku.edu.cn}}

\maketitle              % typeset the header of the contribution
\begin{abstract}
Most of the Chinese pre-trained models adopt characters as basic units for downstream tasks. However, these models ignore the information carried by words and thus lead to the loss of some important semantics.
In this paper, we propose a new method to exploit word structure and integrate lexical semantics into character representations of pre-trained models. Specifically, we project a word's embedding into its internal characters' embeddings according to the similarity weight. To strengthen the word boundary information, we mix the representations of the internal characters within a word. After that, we apply a word-to-character alignment attention mechanism to emphasize important characters by masking unimportant ones.
Moreover, in order to reduce the error propagation caused by word segmentation, we present an ensemble approach to combine segmentation results given by different tokenizers.
The experimental results show that our approach achieves superior performance over the basic pre-trained models BERT, BERT-wwm and ERNIE on different Chinese NLP tasks:
sentiment classification, sentence pair matching, natural language inference and machine reading comprehension. We make further analysis to prove the effectiveness of each component of our model.
\keywords{word semantics  \and character representation \and pre-trained models}
\end{abstract}

\section{Introduction}
Pre-trained language models (PLMs) such as BERT~\cite{devlin2018bert}, RoBERTa~\cite{liu2019roberta} and XLNet~\cite{yang2019xlnet} have shown great power on a variety of natural language processing (NLP) tasks, such as natural language understanding, text classification and automatic question answering. In addition to English, pre-trained models also prove their effectiveness on Chinese NLP tasks~\cite{sun2019ernie,cui2019pre}.

Pre-trained models were originally designed for English, where spaces are considered as natural delimiters between words. But in Chinese, since there is no explicit word boundary, it is intuitive to utilize characters directly to build pre-trained models, such as BERT for Chinese~\cite{devlin2018bert}, BERT-wwm~\cite{cui2019pre} and ERNIE~\cite{sun2019ernie}. In Chinese, words instead of characters are the basic semantic units that have specific meanings and can behave independently~\cite{LNCG} to build a sentence. The meaning of a single Chinese character is always ambiguous.
For example, ``\begin{CJK*}{UTF8}{gbsn}足\end{CJK*}'' has four meanings specified in the Chinese dictionary: foot, attain, satisfy and enough. Therefore, One single character can not express the meaning precisely. In contrast, the Chinese word which is constructed by combining several characters can accurately express semantic meaning. For example, if the character ``\begin{CJK*}{UTF8}{gbsn}足\end{CJK*}'' is combined with the character ``\begin{CJK*}{UTF8}{gbsn}不\end{CJK*}(no)'', we get the Chinese word ``\begin{CJK*}{UTF8}{gbsn}不足(insufficient)\end{CJK*}''. If the character ``\begin{CJK*}{UTF8}{gbsn}足\end{CJK*}'' is combined with the character ``\begin{CJK*}{UTF8}{gbsn}球\end{CJK*}(ball)'', we get the Chinese word ``\begin{CJK*}{UTF8}{gbsn}足球(soccer)\end{CJK*}''. Therefore, to understand Chinese text, the knowledge of words can greatly avoid ambiguity caused by single characters. 

Previous ways to integrate word information into the pre-trained model can be categorized into two ways, as shown in Figure 1. The first way (shown as 1(a)) is to splice character information and word information, and merge them through the self-attention mechanism~\cite{mengge2020porous,li2020flat,lai2021lattice}. This type of method increases the length of the input sequence, and because the self-attention mechanism is position-insensitive, the model needs to be specified with the positional relation between characters and their corresponding words. The other way (shown as 1(b)) is to add word information into their internal characters' information, which can either use one encoder to operate~\cite{liu2021lexicon}, or use two encoders to encode characters and words separately~\cite{diao2019zen}. 
Previous work tends to integrate multiple words' information into a character representation, which might introduce redundant information and bring noises. Our work is in line with Method (b), but we design a more comprehensive strategy to exploit word semantics and enrich character representation.

\begin{figure}[tbp]
\centering
\includegraphics[width=8cm]{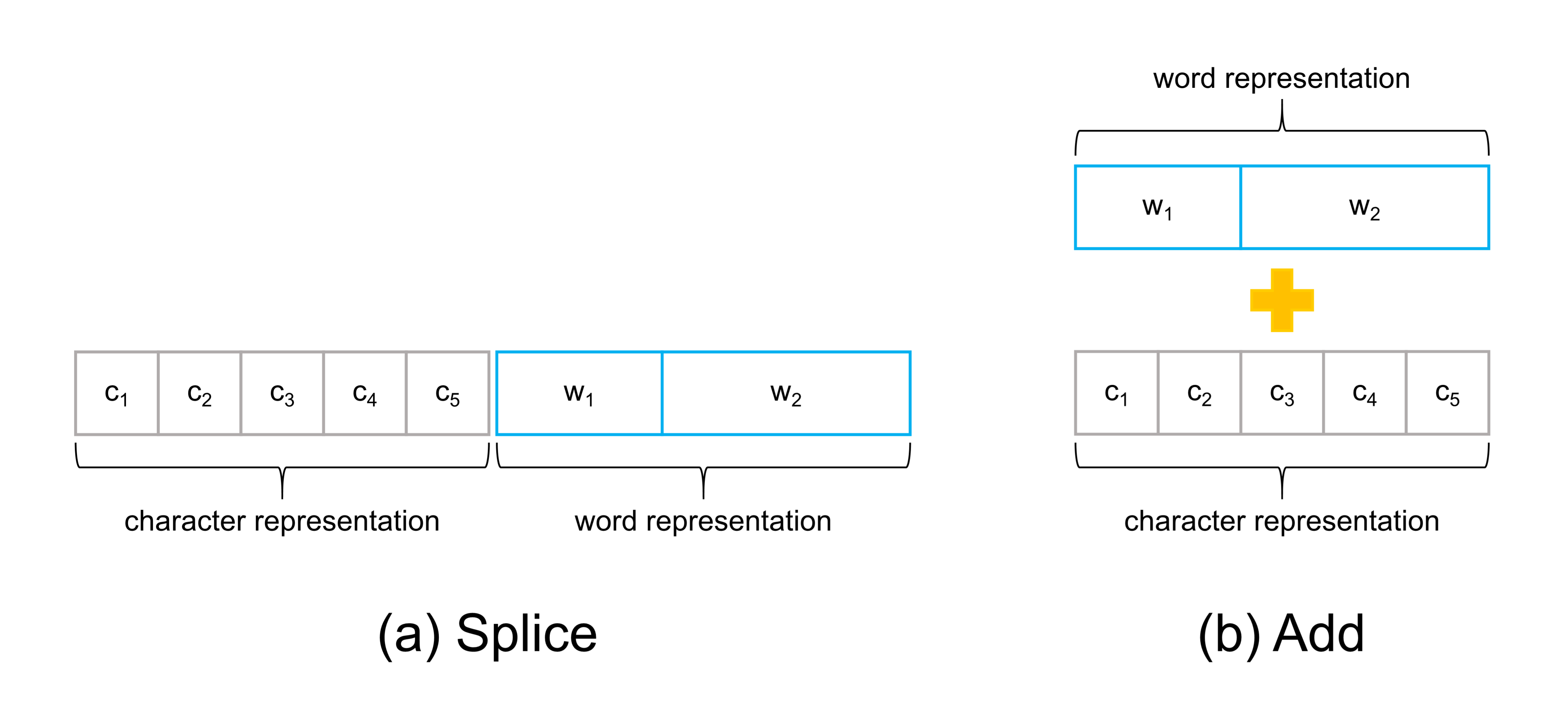}
\caption{Two ways to integrate word information to character representation.}
\end{figure}

% The Chinese pre-trained models, such as BERT-wwm~\cite{cui2019pre}, ERNIE \cite{sun2019ernie}, ZEN \cite{diao2019zen}, Lattice-BERT ~\cite{lai2021lattice}, try to employ word information to guide pre-training, where the two methods shown in Figure 1 are utilized to facilitate the fusion of word-level and character-level knowledge. But at the user side, the interface for downstream tasks only has character features, thus the inconsistency between the pre-training and fine-tuning stages will cause certain deviations.

% In the fine-tuning stage, several approaches are proposed to employ word information to enhance pre-trained models. \cite{liu2021lexicon} combine lexicon information and pre-trained models to explore Chinese sequence labeling tasks. \cite{li2019enhancing} propose a Multi-source Word Aligned Attention (MWA) mechanism, where they take advantage of word boundary information and 
% allocate the same proportion of attention to each intra-word character. This work only considers the shallow word boundary information but ignores the deep word semantics, and moreover, this work ignores the fact that the importance of a character in different words varies according to contexts.

In this paper, we propose a new method, Hidden Representation Mix and Fusion (HRMF), exploiting word semantics to enrich character representations of Chinese pre-trained models in the fine-tuning stage. Firstly, the importance of each character in a word is calculated based on the cosine similarity between the character's representation and its paired word's representation. Then we integrate the word embedding into the representation of each internal character according to the importance weight. After that, we apply a mixing mechanism to enable characters within a word exchange information with each other to further enhance the character representation. In addition, we apply a multi-head attention and a masked multi-head attention which masks the unimportant characters, and the final representation is obtained by fusing the representations given by the two attention mechanisms, which is then applied in downstream tasks. Moreover, in order to minimize the impact of word segmentation errors, we adopt a multi-tokenizer voting mechanism to obtain the final word segmentation result. 

We conduct extensive experiments on several NLP tasks, including sentiment classification, sentence pair matching, natural language inference and machine reading comprehension. Experimental results show that our proposed method consistently improves the performance of BERT, BERT-wwm and ERNIE on different NLP tasks, and the ablation study shows that each component of our model yields improvement over the basic pre-trained models. Our code are made available at \href{https://github.com/liwb1219/HRMF}{https://github.com/liwb1219/HRMF}.
% We will make our code public available at GitHub for further research. 

To sum, the contributions of this paper are as follows:
\begin{itemize}
\item We define the character's  importance in a word, which enables our model to lay particular emphasis on important characters in a word. 
\item We apply a mixing mechanism to facilitate information exchange between characters, which further enriches our character representation.
\item We present a masked multi-head attention mechanism to discard the semantics of unimportant characters in a word, which provides a supplement to the original sentence representation.
\item  Our proposed method outperforms the main-stream Chinese pre-trained models on a variety of NLP tasks.
\end{itemize}

\section{Related Work}
To exploit word information and help the model extract a enhanced representation of Chinese characters,
the related work can be roughly divided into traditional methods and transformer-based methods.

In traditional methods, Su~\etal~\cite{su2017lattice} propose a word-lattice based Recurrent Neural Network (RNN) encoder for NMT, which generalizes the standard RNN to a word lattice topology. Zhang~\etal~\cite{zhang2018chinese} propose a word-lattice based LSTM network to integrate latent word information into the character-based LSTM-CRF model, which uses gated cells to dynamically route information from different paths to each character. Ma~\etal~\cite{ma2019simplify} propose a simple but effective method for incorporating lexicon information into the character representations, which requires only a subtle adjustment of the character representation layer to introduce the lexicon information.

As for the transformer-based pre-trained method, there are two main approaches: Splice and Add, as illustrated in Figure 1.

\textbf{Splice}. Pre-trained models are mostly constructed through the multi-head self-attention mechanism, which is not sensitive to location, the location information needs to be clearly specified in the model. Xue~\etal~\cite{mengge2020porous} propose PLTE, which models all the characters and matched lexical words in parallel with batch processing. Lai~\etal~\cite{lai2021lattice} propose a novel pre-training paradigm for Chinese—Lattice-BERT, which explicitly incorporates word representations along with characters, thus can model a sentence in a multi-granularity manner.  Li~\etal~\cite{li2020flat} propose FLAT: Flat-LAttice Transformer for Chinese NER, which converts the lattice structure into a flat structure consisting of spans. Each span corresponds to a character or latent word and its position in the original lattice.

\textbf{Add}. The Splice method not only increases the complexity of sequence modeling, but also needs to be specified with the positional relation between characters and words. The Add method is relatively flexible, where the word information can be added to it's corresponding character representation based on word boundary. The model can use one encoder or two. By using one encoder, the encoder can be used to simultaneously encode character information and word information. As for two encoders, they can be used to encode character information and word information separately. Diao~\etal~\cite{diao2019zen} propose ZEN, a BERT-based Chinese text encoder enhanced by N-gram representations. They use one encoder to encode character and the other encoder to encode N-gram, and combine the two lines of information at each layer of the model. Liu~\etal~\cite{liu2021lexicon} propose Lexicon Enhanced BERT (LEBERT) to integrate lexicon information between Transformer layers of BERT directly, where a lexicon adapter is designed to dynamically extract the most relevant matched words for each character using a char-to-word bilinear attention mechanism, and then is applied between adjacent layers.

% In addition, \cite{meng2019glyce} propose the GLYCE, treating Chinese characters as images and use CNNs to obtain their representations. \cite{sun2021chinesebert} propose ChineseBERT, which integrates both the glyph and pinyin information of Chinese characters into language model pre-training. \cite{tian2020joint} propose a character-based neural model for the joint task enhanced by multi-channel attention of n-grams. In the attention module, n-gram features are categorized into different groups according to several criteria, and n-grams in each group are weighted and distinguished according to their importance for the joint task in the specific context.

\section{Multiple Word Segmentation Aggregation}
% Word segmentation is the basis of Chinese language processing. Errors in word segmentation will lead to a drop in performance of upper-level applications. In order to improve the robustness of the word segmentation result, we adopt an ensemble strategy, using multiple tokenizers to vote on the word segmentation results. 

In the experiment, we use three tokenizers to vote, namely PKUSEG~\cite{luo2019pkuseg}, LAC~\cite{jiao2018chinese} and snownlp\footnote{\url{https://github.com/isnowfy/snownlp}}.
With the segmentation results given by the tokenizers, we select the final segmentation result following two rules, the majority rule and the granularity rule. Firstly, with the beginning character $c_s$, we extract the segmented word starts with $c_s$ from the segmentation results of different tokenizers. We get $\{T^A_{c_s},T^B_{c_s},T^C_{c_s}\}$, which means the set of segmented words from different tokenizers starting with $c_s$. Then, following the majority rule, we select the one appears the most in $\{T^A_{c_s},T^B_{c_s},T^C_{c_s}\}$ as the first part of the segmentation result. If there are two words having the same time of occurrence, we choose the one with larger granularity. After determining the first segmented word $T_{c_s}$, we set the character after $T_{c_s}$ as the next beginning character and repeat this process until we get the final segmentation result. For a clear understanding, Figure 2 shows an example.

\begin{figure}[htbp]
\centering
\includegraphics[width=6cm]{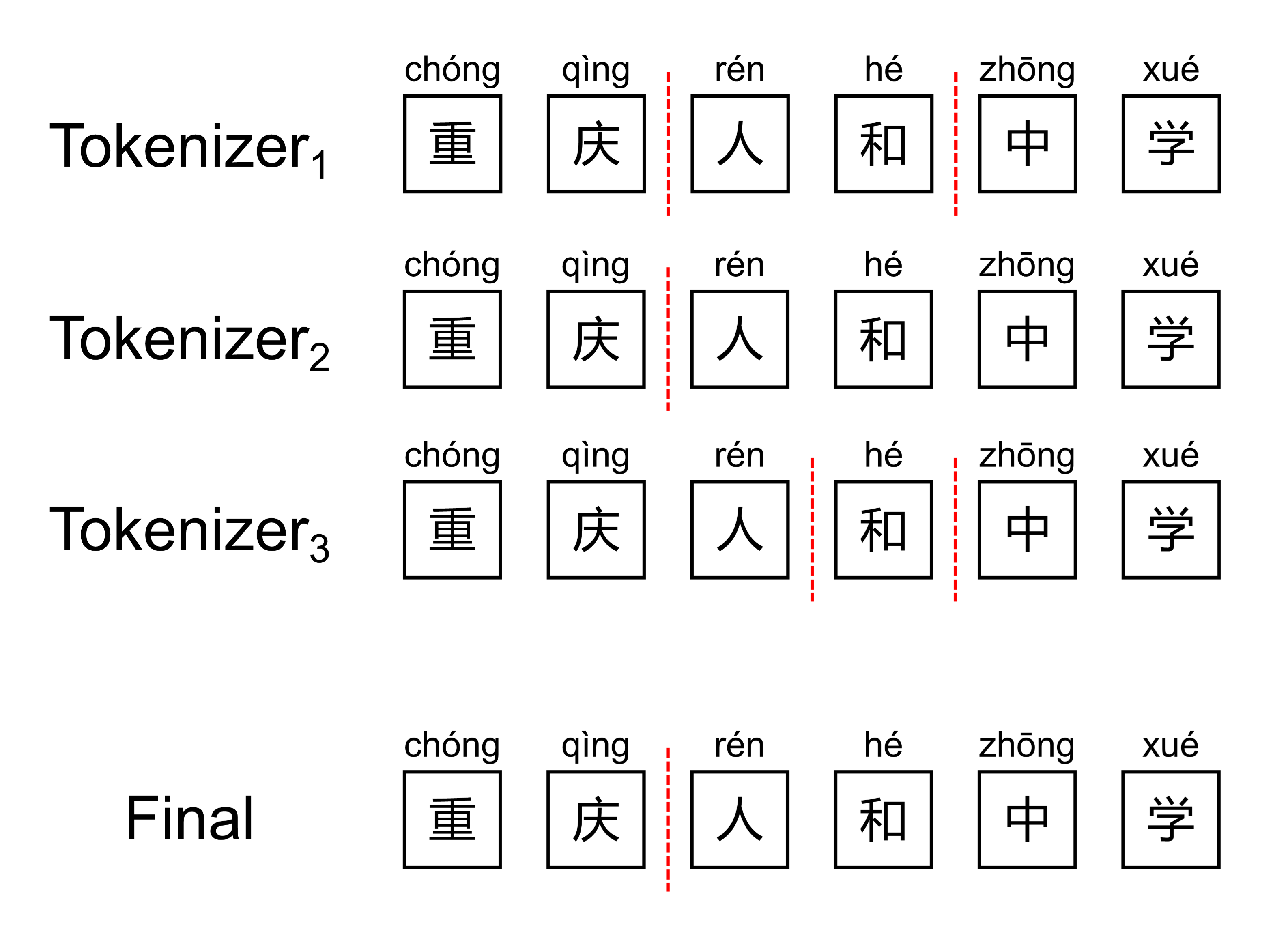}
\caption{Example of voting segmentation plan. Firstly, with the character ``\begin{CJK*}{UTF8}{gbsn}重\end{CJK*}'' as the beginning character, ``\begin{CJK*}{UTF8}{gbsn}重庆\end{CJK*}'' appears twice, and ``\begin{CJK*}{UTF8}{gbsn}重庆人\end{CJK*}'' appears once. Following the majority rule, we choose ``\begin{CJK*}{UTF8}{gbsn}重庆\end{CJK*}'' as the first part of the segmentation result. Then, we choose ``\begin{CJK*}{UTF8}{gbsn}人\end{CJK*}'' as the beginning character. ``\begin{CJK*}{UTF8}{gbsn}人和\end{CJK*}'' appears once, and ``\begin{CJK*}{UTF8}{gbsn}人和中学\end{CJK*}'' appears once, which has the same time of occurrence. Following the granularity rule, the word with larger granularity is preferred, so we choose ``\begin{CJK*}{UTF8}{gbsn}人和中学\end{CJK*}''. Finally, the word segmentation result is ``\begin{CJK*}{UTF8}{gbsn}重庆/人和中学\end{CJK*}''.}
\end{figure}

\section{Projecting Word Semantics to Character Representation}
The overall structure of our model is shown in Figure 3. We employ the pre-trained model BERT and its updated variants (BERT-wwm, ERNIE) to obtain the original character representation, and a parallel multi-head attention module is applied to facilitate the fusion of character representation and word semantics.

\begin{figure}[htbp]
\centering
\includegraphics[width=12cm]{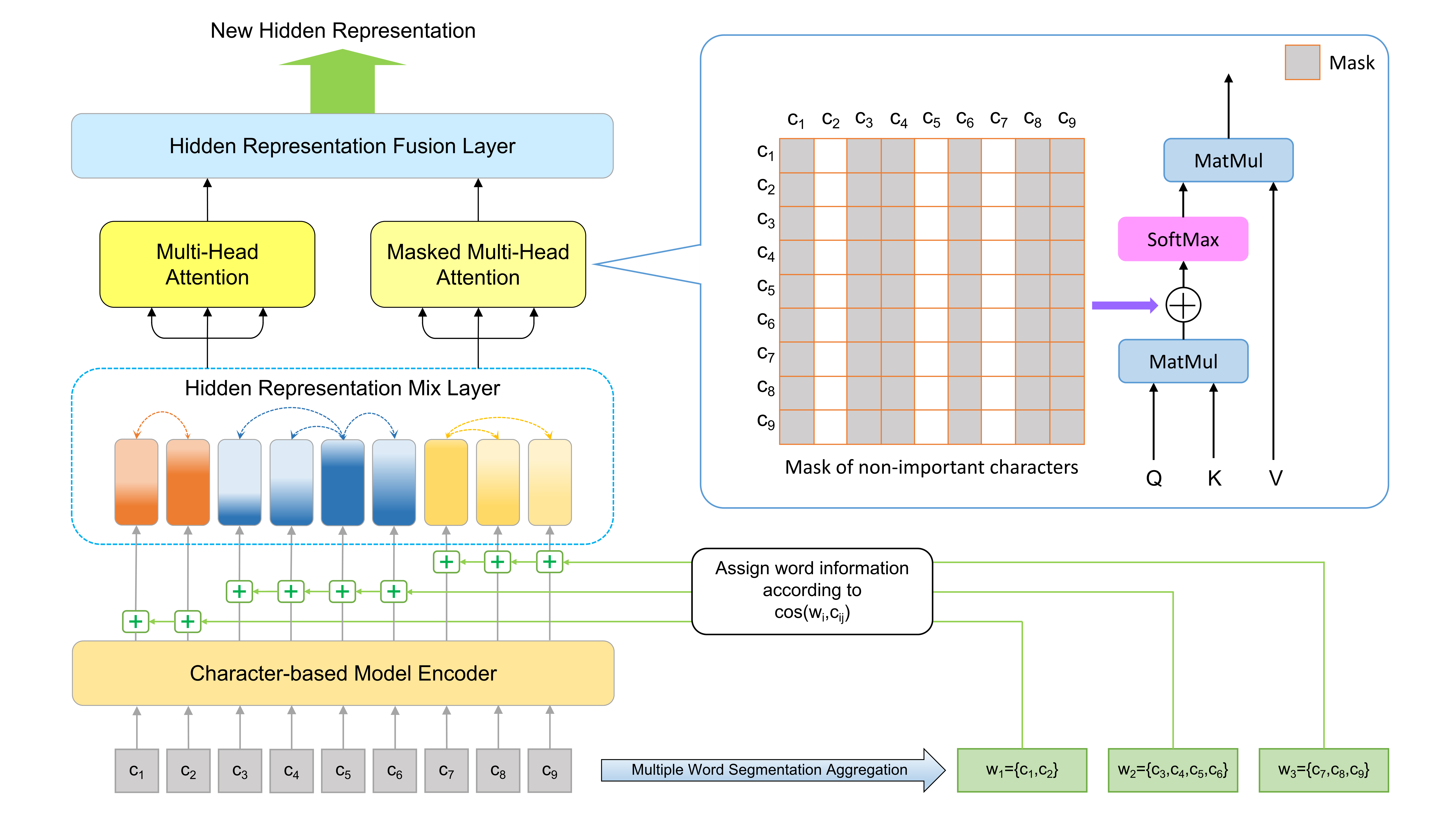}
\caption{The architecture of our proposed model. The color shades of squares in the hidden representation mix layer indicate the importance weights of different characters. The masked multi-head attention means that unimportant characters are masked when calculating the attention score.}
\end{figure}

\subsection{Integrating Word Embedding to Character Representation}
At present, the mainstream Chinese pre-trained models usually use character-level features. In order to make use of lexical semantics, we integrate word knowledge into character representation. Specifically, we first calculate the cosine similarity between the character and its corresponding word. And then, we assign the word information to its internal characters based on the similarity score. 

The input text is denoted as $\mathbf{S}=(c_1, c_2,..., c_n)$, and the hidden layer representation after BERT encoding is expressed as $\mathbf{H}=\text{BERT}(\mathbf{S})=(h_1, h_2,..., h_n)$. The $t$-th word in the text is 
$w_t=(c_i, ..., c_j)$. We first obtain word embedding by:
\begin{equation}
x_t=e^w(w_t)
\end{equation}
where $e^w$ is a pre-trained word embedding lookup table \cite{song2018directional}. 

We use two linear layers to transform dimensions and learn the difference between two different sets of vector spaces.
% We perform the dimension conversion operation by: 

\begin{equation}
v_t=(\text{tanh}(x_t\mathbf W_1+ \mathbf b_1))\mathbf W_2+ \mathbf b_2
\end{equation}
where $\mathbf W_1 \in \mathbb{R}^{d_w \times d_h}$, $\mathbf W_2 \in \mathbb{R}^{d_h \times d_h}$, $\mathbf b_1$ and $\mathbf b_2$ are scaler bias. $d_w$ and $d_h$ denote the dimension of word embedding and the hidden size of BERT respectively.

To measure the semantic weight of each character within a word, we compute the cosine similarity: 

\begin{equation}
\text{score}_{k}=\text{cos}(h_k,v_t),k=i,...,j
\end{equation}
where $\text{score}_{k}$ represents the cosine similarity between the word and its $k$-th character.

We integrate word information into character embeddings based on the similarity score: 
\begin{equation}
h^w_k=h_k+\frac{\text{score}_k}{\sum\limits_{m=i}^j \text{score}_m}v_t,k=i,...,j
\end{equation}
where $h_k^w$ denotes the updated hidden representation of character $k$.

\subsection{Mixing Character Representations within a Word}
In order to enrich the representations of characters, we mix the embeddings of characters within a word, that is, we let the characters in one word exchange information with each other. First, we define the key character as the one that has the highest similarity score with the corresponding word:
\begin{equation}
p=\text{argmax}(\text{score}_k),k=i,...,j
\end{equation}
where $p$ denotes the index of the key character in a word.

Accordingly, the key character collect information from all other characters, and at the same time, the key character also passes its own information to all other characters.

Specifically, for single-character words, they keep the original representations. For two-character words, we let the key character passes a small amount of information to the non-important character, and the non-important character  gives a small amount of information to the key character. For multi-character words (larger than or equal to three), we let the key character passes a small amount of information to all non-important characters, and all non-important characters give a small amount of information to the key character, while there is no information exchange among non-important characters.

We set a parameter $\lambda$ as the retention ratio of the key character information.We introduce a non-linear function to degrade its reduction rate. The hidden representation of characters in non-single-character words is calculated as follows:

\begin{equation}
\widetilde{h_k^w}=\begin{cases}
f(\lambda)h_k^w+g(\lambda)\sum\limits_{\substack{i \leq m \leq j \\ m\neq p}}h_m^w,&k=p\\
\\
g(\lambda)h_p^w+(1-g(\lambda))h_k^w,&k\neq p
\end{cases}
\end{equation}

\begin{equation}
f(\lambda)=e^{\lambda-1} 
\end{equation}

\begin{equation}
g(\lambda)=\frac{1-f(\lambda)}{j-i}
\end{equation}
where $\widetilde{h_k^w}$ is the mixed hidden representation of character $k$.

Therefore, the new sequence is represented as:
\begin{equation}
\mathbf{\widetilde{H}}=(\widetilde{h_1^w},\widetilde{h_2^w},...,\widetilde{h_n^w})
\end{equation}

\subsection{Fusing New Character Embedding to Sentence Representation}
We apply the self-attention mechanism on the mixed hidden representation obtained in the previous step to get a new hidden representation $\mathbf{H^1}$:
\begin{equation}
\mathbf{H^1}=\text{softmax}(\frac{(Q\mathbf W_q^1)(K\mathbf W_k^1)}{\sqrt{d_h}})(V\mathbf W_v^1)
\end{equation}
where $Q$, $K$ and $V$ are all equal to the collective representation $\mathbf{\widetilde{H}}$ obtained in the previous step. $\mathbf W_q^1$, $\mathbf W_k^1$, $\mathbf W_v^1$ are trainable parameter matrices.

Taking into account that the non-important characters' information has been integrated into the key character's representation, when calculating the attention score, those non-important characters can be masked. In this way, a new hidden representation $\mathbf H^2$ is obtained by a masked self-attention: 

\begin{equation}
\mathbf{H^2}=\text{softmax}(\frac{(Q\mathbf W_q^2)(K\mathbf W_k^2)}{\sqrt{d_h}}+mask)(V\mathbf W_v^2)
\end{equation}

\begin{equation}
mask_{ij}=\begin{cases}
0,j \in \mathbf{\Omega} \\
\\
-inf,j \notin \mathbf{\Omega}
\end{cases}
\end{equation}
where $\mathbf W_q^2$, $\mathbf W_k^2$, $\mathbf W_v^2$ are trainable parameter matrices, $mask \in \mathbb{R}^{n \times n}$, $\mathbf{\Omega}$ is the set of subscripts corresponding to important words.

The final representation is obtained by fusing two sorts of embeddings:

\begin{equation}
\mathbf{\hat{H}}=\mu \mathbf{H^1}+(1-\mu)\mathbf{H^2}
\end{equation}
where $\mathbf{\hat{H}}$ is the final hidden representation, which will be applied to downstream tasks.

\section{Experimental Setup}
\subsection{Tasks and Datasets}
In order to prove the effectiveness of our proposed method, 
% we chose three encoders(BERT, BERT-wwm or ERNIE) to encode character features, and 
we conduct experiments on the following four public datasets with several NLP tasks. For data statistics of these datasets, please refer to Table 1.

\textbf{Sentiment Classification (SC)}: ChnSentiCorp\footnote{\url{https://github.com/pengming617/bert_classification}} is a Chinese sentiment analysis data set, containing online shopping reviews of hotels, laptops and books. 
% The data set is automatically collected from the Ctrip website and sorted out. The size of the corpus is about 10,000 pieces.

\textbf{Sentence Pair Matching (SPM)}: LCQMC~\cite{liu2018lcqmc} is a large Chinese question matching corpus, aiming to identify whether two sentences have the same meaning. 
% It is more challenging than paraphrase corpus as it focuses on intent matching rather than paraphrase. Each sample in the data set is associated with a pair of sentences and a binary label, indicating whether the two sentences have the same intent. The task can be formalized as predicting a binary label "Yes/No".

\textbf{Natural Language Inference (NLI)}: XNLI~\cite{conneau2018xnli} is a cross-lingual natural language inference corpus, which is a crowdsourced collection of multilingual corpora. 
% These pairs have textual implication notes and have been translated into 14 languages including Chinese. The label contains "contradiction, neutral and entailment". 
We only use the Chinese part.

\textbf{Machine Reading Comprehension (MRC)}: DRCD~\cite{shao2018drcd} is a span-extraction MRC dataset written in Traditional Chinese.
% MRC is a representative document-level modeling task which requires to answer questions based on the given passage. We select DRCD~\cite{shao2018drcd} data, which is a span-extraction MRC dataset written in Traditional Chinese.

\begin{table}[htbp]
\caption{Hyper-parameter settings and data statistics in different tasks. Blr* represents the initial learning rate of BERT/BERT-wwm model for the AdamW optimizer.}
\centering
\resizebox{\textwidth}{12mm}{
\begin{tabular}{@{}ccccccccccc@{}}
\hline
Dataset & Task & MaxLen & Batch & Epoch & Blr* & ERNIE lr* & Train & Dev & Test & Domain \\
\hline
ChnSentiCorp & SC & 256 & 64 & 4 & 3e-5 & 3e-5 & 9.6K & 1.2K & 1.2K & various \\
LCQMC & SPM & 128 & 64 & 3 & 2e-5 & 3e-5 & 239K & 8.8K & 12.5K & Zhidao \\
XNLI & NLI & 128 & 64 & 2 & 3e-5 & 5e-5 & 393K & 2.5K & 5K & various \\
DRCD & MRC & 512 & 16 & 2 & 3e-5 & 5e-5 & 27K & 3.5K & 3.5K & Wikipedia \\ 
\hline
\end{tabular}}
\end{table}

\subsection{Baseline Models}
We adopt the pre-trained models BERT~\cite{devlin2018bert}, BERT-wwm~\cite{cui2019pre} and ERNIE~\cite{sun2019ernie} as our base architectures.

% \textbf{BERT}~\cite{devlin2018bert}. BERT is designed to pre-train deep bidirectional representations of unlabeled text by jointly conditioning on both left and right context. 

% \textbf{BERT-wwm}~\cite{cui2019pre}. The BERT-wwm model adopts the whole word masking strategy on Chinese BERT by masking all the Chinese characters in the same word. In this way, the model can learn some semantic information of words.

% \textbf{ERNIE}~\cite{sun2019ernie}. In addition to basic masking strategy, the ERNIE model uses two kinds of knowledge strategies: phrase-level strategy and entity-level strategy. It takes a phrase or an entity as one unit, and all of the words in the same unit are masked during word representation training.

\subsection{Training Details}
In order to ensure the fairness and robustness of the experiment, for the same dataset and encoder, we use the same parameters, such as maximum length, warm-up ratio, initial learning rate, optimizer, etc. We repeated each experiment five times and reported the average score.

We do experiments using the Pytorch~\cite{paszke2019pytorch} framework, and all the baseline weight files were converted to the Pytorch version. At training time, we fix the pretrained word embeddings, using the AdamW  optimizer~\cite{loshchilov2018fixing}, the weight decay is 0.02, and the warm-up ratio is 0.1. The proportion of the key character information retention $\lambda$ is set to 0.9. The fusion coefficient $\mu$ is set to 0.5. For detailed hyper-parameter settings, please see Table 1.

\section{Results and Analysis}
\subsection{Overall Results}
Table 2 shows the experimental results on four public data sets with different NLP tasks, which demonstrates that our method obtains an obvious improvement compared with the baseline pre-trained models.

\begin{table}[htbp]
\caption{Experimental results on four datasets. The classification tasks (SC, SPM, NLI ) adopt $Accuracy$ as the evaluation metric, and the machine reading comprehension task adopts $EM$ and $F1$ as evaluation metrics.}
\centering
\begin{tabular}{c|c|c|c|c|c|c}
\hline
Task & SC & SPM & NLI & MRC & \multicolumn{2}{c}{avg.} \\
\hline
Dataset & ChnSentiCorp & LCQMC & XNLI  & DRCD[EM / F1] & cls & all \\
\hline
BERT              & 94.72 & 86.61 & 77.85 & 85.48 / 91.36 & 86.39  & 87.20      \\
+HRMF             & 95.48 & 87.24 & 78.44 & 86.32 / 91.76 & 87.05  & 87.85      \\ \hline
BERT-wwm          & 94.82 & 86.67 & 78.02 & 85.79 / 91.66 & 86.50  & 87.39      \\
+HRMF             & 95.36 & 86.96 & 78.21 & 86.55 / 92.11 & 86.84  & 87.84      \\ \hline
ERNIE             & 95.32 & 87.26 & 78.26 & 87.78 / 93.20 & 86.95  & 88.36      \\
% ERNIE+HRMF        & 96.12 & 88.29 & 78.87 & 88.59 & 93.70 & 87.76  & 89.11      \\ \hline
+HRMF             & \textbf{96.12} & \textbf{88.29} & \textbf{78.87} & \textbf{88.59} / \textbf{93.70} & \textbf{87.76} & \textbf{89.11} \\ \hline
\end{tabular}
\end{table}

Specifically, in the classification task, our method yields the most obvious improvement in sentiment analysis, with an improvement of 0.76 over BERT, 0.54 over BERT-wwm and 0.80 over ERNIE.
We believe that this is because the emotional polarity of a sentence is more sensitive to word semantics. In other words, the emotional tendency of a sentence is likely to be determined by some of the words. Similarly, in sentence meaning matching and natural language inference, our method also achieves an average improvement of 0.65 and 0.46 compared with three baseline models. As for the task of machine reading comprehension, our method significantly improves the EM index with an average improvement of 0.80, which shows that after incorporating word knowledge, the model judges the boundary of answers more accurately. 
%Since F1 is a loose metric on this task compared to EM, 
The baseline models already achieve a relatively high F1 score, and our method still obtains an average improvement of 0.45 on F1.

\subsection{Ablation Study}
In order to verify the effectiveness of our method, we strip off different parts of the model to conduct experiments, as shown in Table 3.

\begin{table}[htbp]
\caption{Ablation study of different components. As shown in Figure 2, $HRMF$ is the complete model of this paper. $-HRML$ is to remove the hidden representation of mix layer; $-MMHA$ means to remove the masked multi-head attention module.}
\centering
\begin{tabular}{c|c|c|c|c}
\hline
Task    & SC           & SPM   & NLI   & MRC             \\ \hline
Dataset & ChnSentiCorp & LCQMC & XNLI  & DRCD[EM / F1] \\ \hline
BERT        & 94.72        & 86.61 & 77.85 & 85.48 / 91.36            \\
BERT+HRMF       & 95.48        & 87.24 & 78.44 & 86.32 / 91.76            \\
BERT+HRMF-HRML       & 95.30        & 86.92 & 78.18 & 86.00 / 91.73       \\
BERT+HRMF-MMHA       & 95.40        & 87.14 & 78.16 & 86.09 / 91.77       \\  \hline
\end{tabular}
\end{table}

After stripping off the hidden representation mix layer, the model has an obvious drop in the performance. It decreases by 0.18, 0.32, 0.26, 0.32 respectively on the four data sets. The performance of the model declines the most compared with other stripped off models on three data sets, which shows that mixing the representation of the characters within a word greatly enhances the representation of the original character representation. 
Besides, after stripping off the masked multi-head attention module, the model also gets a drop in the performance, which demonstrates that downstream tasks benefit from the additional representation of masked multi-head attention module that masks unimportant characters.

\begin{figure}[htb]
\centering
\includegraphics[width=12cm]{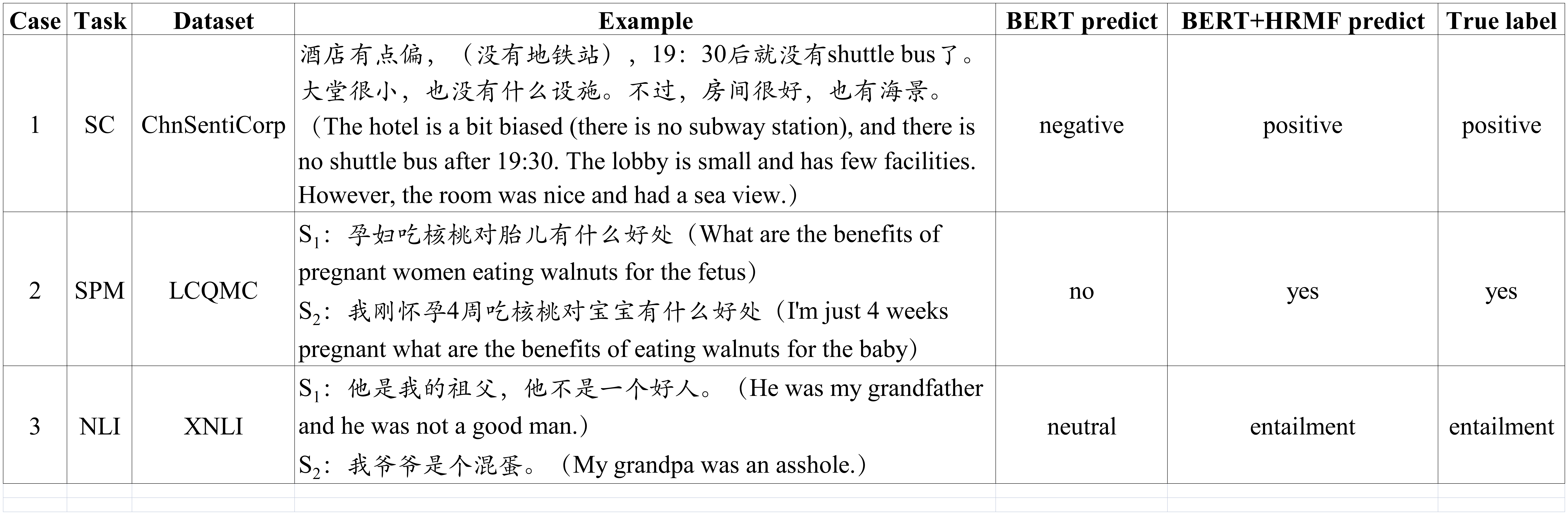}
\caption{Some examples of classification tasks.}
\end{figure}
\subsection{Case Study}
To better demonstrate our model's improvement on the understanding of semantics, we conduct case studies on several specific instances, as shown in Figure 4 for classification tasks.

Compared with BERT, our proposed model obtains better performance on several tasks. On the Sentiment Classification (SC) task, our model has a better understanding on the sentences that have a emotional semantic transition. For example, in Case 1, the sentences' emotional tendencies change from negative to positive through the word "\begin{CJK*}{UTF8}{gbsn}不过\end{CJK*}(However)", which is accurately understood by our model. On the Sentence Pair Matching (SPM) task, our proposed model shows a better understanding on the relation between sentences' semantic information. For example, our model can accurately understand that "\begin{CJK*}{UTF8}{gbsn}胎儿\end{CJK*}(fetus)" has the same meaning from "\begin{CJK*}{UTF8}{gbsn}宝宝\end{CJK*}(baby)" in Case 2. On the Natural Language Inference (NLI) task, our model can accurately identify the relation between semantic information and make correct inference. For example, in case 3, our model recognizes that "\begin{CJK*}{UTF8}{gbsn}不是一个好人\end{CJK*}(not a good man)" has the same meaning with "\begin{CJK*}{UTF8}{gbsn}是个混蛋\end{CJK*}(an asshole)". One possible reason that our model outperforms BERT is that our model integrates word semantic information. 

\section{Conclusion}
In this paper, we propose a method HRMF to improve charater-based Chinese pre-trained models by integrating lexical semantics into character representations. We enrich the representations by mixing the intra-word characters' embeddings, and add a masked multi-head attention module by masking unimportant characters to provide a supplement to the original sentence representation. We conduct extensive experiments on four different NLP tasks. %Compared with BERT, BERT-wwm, and ERNIE fine-tuning, 
Based on the main-stream Chinese pre-trained models BERT, BERT-wwm, and ERNIE, our proposed method achieves obvious improvements, which proves its effectiveness and universality. In future work, we will combine more knowledge with Chinese characteristics to further improve Chinese pre-trained models for downstream tasks.

\vspace{10pt}

\noindent\textbf{Acknowledgement.} This work is supported by the National Hi-Tech RD Program of China (2020AAA0106600), the National Natural Science Foundation of China (62076008) and the KeyProject of Natural Science Foundation of China (61936012).

\clearpage
\bibliography{sample}

\begin{thebibliography}{10}
\providecommand{\url}[1]{\texttt{#1}}
\providecommand{\urlprefix}{URL }
\providecommand{\doi}[1]{https://doi.org/#1}

\bibitem{conneau2018xnli}
Conneau, A., Lample, G., Rinott, R., Williams, A., Bowman, S.R., Schwenk, H.,
  Stoyanov, V.: Xnli: Evaluating cross-lingual sentence representations. In
  EMNLP, pp. 2475--2485 (2018)

\bibitem{cui2019pre}
Cui, Y., Che, W., Liu, T., Qin, B., Yang, Z., Wang, S., Hu, G.: Pre-training
  with whole word masking for chinese bert. arXiv preprint arXiv:1906.08101
  (2019)

\bibitem{devlin2018bert}
Devlin, J., Chang, M.W., Lee, K., Toutanova, K.: Bert: Pre-training of deep
  bidirectional transformers for language understanding. In NAACL, pp.
  4171--4186 (2019)

\bibitem{diao2019zen}
Diao, S., Bai, J., Song, Y., Zhang, T., Wang, Y.: Zen: Pre-training chinese
  text encoder enhanced by n-gram representations. In EMNLP, pp. 4729--4740
  (2019)

\bibitem{jiao2018chinese}
Jiao, Z., Sun, S., Sun, K.: Chinese lexical analysis with deep bi-gru-crf
  network. arXiv preprint arXiv:1807.01882  (2018)

\bibitem{lai2021lattice}
Lai, Y., Liu, Y., Feng, Y., Huang, S., Zhao, D.: Lattice-bert: Leveraging
  multi-granularity representations in chinese pre-trained language models. In
  NAACL, pp. 1716--1731 (2021)

\bibitem{li2020flat}
Li, X., Yan, H., Qiu, X., Huang, X.: Flat: Chinese ner using flat-lattice
  transformer. In ACL,  (2020)

\bibitem{liu2021lexicon}
Liu, W., Fu, X., Zhang, Y., Xiao, W.: Lexicon enhanced chinese sequence
  labelling using bert adapter. In ACL, pp. 5847--5858 (2021)

\bibitem{liu2018lcqmc}
Liu, X., Chen, Q., Deng, C., Zeng, H., Chen, J., Li, D., Tang, B.: Lcqmc: A
  large-scale chinese question matching corpus. In COLING, pp. 1952--1962
  (2018)

\bibitem{liu2019roberta}
Liu, Y., Ott, M., Goyal, N., Du, J., Joshi, M., Chen, D., Levy, O., Lewis, M.,
  Zettlemoyer, L., Stoyanov, V.: Roberta: A robustly optimized bert pretraining
  approach. arXiv preprint arXiv:1907.11692  (2019)

\bibitem{loshchilov2018fixing}
Loshchilov, I., Hutter, F.: Fixing weight decay regularization in adam  (2018)

\bibitem{luo2019pkuseg}
Luo, R., Xu, J., Zhang, Y., Ren, X., Sun, X.: Pkuseg: A toolkit for
  multi-domain chinese word segmentation. CoRR  \textbf{abs/1906.11455} (2019)

\bibitem{ma2019simplify}
Ma, R., Peng, M., Zhang, Q., Huang, X.: Simplify the usage of lexicon in
  chinese ner. In ACL, pp. 5951--5960 (2019)

\bibitem{mengge2020porous}
Mengge, X., Yu, B., Liu, T., Zhang, Y., Meng, E., Wang, B.: Porous lattice
  transformer encoder for chinese ner. In COLING, pp. 3831--3841 (2020)

\bibitem{paszke2019pytorch}
Paszke, A., Gross, S., Massa, F., Lerer, A., Bradbury, J., Chanan, G., Killeen,
  T., Lin, Z., Gimelshein, N., Antiga, L., et~al.: Pytorch: An imperative
  style, high-performance deep learning library. In NeurIPS, pp. 8026--8037
  (2019)

\bibitem{shao2018drcd}
Shao, C.C., Liu, T., Lai, Y., Tseng, Y., Tsai, S.: Drcd: a chinese machine
  reading comprehension dataset. arXiv preprint arXiv:1806.00920  (2018)

\bibitem{song2018directional}
Song, Y., Shi, S., Li, J., Zhang, H.: Directional skip-gram: Explicitly
  distinguishing left and right context for word embeddings. In NAACL, pp.
  175--180 (2018)

\bibitem{su2017lattice}
Su, J., Tan, Z., Xiong, D., Ji, R., Shi, X., Liu, Y.: Lattice-based recurrent
  neural network encoders for neural machine translation. In AAAI, pp.
  3302--3308 (2017)

\bibitem{sun2019ernie}
Sun, Y., Wang, S., Li, Y., Feng, S., Chen, X., Zhang, H., Tian, X., Zhu, D.,
  Tian, H., Wu, H.: Ernie: Enhanced representation through knowledge
  integration. arXiv preprint arXiv:1904.09223  (2019)

\bibitem{yang2019xlnet}
Yang, Z., Dai, Z., Yang, Y., Carbonell, J., Salakhutdinov, R.R., Le, Q.V.:
  Xlnet: Generalized autoregressive pretraining for language understanding. In
  NeurIPS, pp. 5754--5764 (2019)

\bibitem{zhang2018chinese}
Zhang, Y., Yang, J.: Chinese ner using lattice lstm. In ACL, pp. 1554--1564
  (2018)

\bibitem{LNCG}
Zhu, D.: Lexical Notes on Chinese Grammar(in Chinese). The Commercial Press
  (1982)

\end{thebibliography}
\end{document}